\documentclass{article}

\usepackage[english]{babel}
\usepackage[letterpaper,top=2cm,bottom=2cm,left=3cm,right=3cm,marginparwidth=1.75cm]{geometry}
\usepackage{placeins}
\usepackage{float}
\usepackage{svg}
\usepackage{amsmath}
\usepackage{wrapfig}
\usepackage{graphicx}
\usepackage{multirow}
\usepackage{array}
\usepackage[colorlinks=true, allcolors=blue]{hyperref}

\usepackage[sorting=none, style=nature]{biblatex}

\def\tabindent{4mm}
\title{Leveraging Computer Vision in the Intensive Care Unit (ICU) for Examining Visitation and Mobility}
\author{Scott Siegel 
\and{Jiaqing Zhang}
\and{Sabyasachi Bandyopadhyay}
\and{Subhash Nerella}
\and{Brandon Silva}
\and{Tezcan Baslanti}
\and{Azra Bihorac}
\and{Parisa Rashidi}}
 
\begin{document}
\maketitle

\begin{abstract}

Despite the importance of closely monitoring patients in the Intensive Care Unit (ICU), many aspects are still assessed in a limited manner due to the time constraints imposed on healthcare providers. For example, although excessive visitations during rest hours can potentially exacerbate the risk of circadian rhythm disruption and delirium, it is not captured in the ICU. Likewise, while mobility can be an important indicator of recovery or deterioration in ICU patients, it is only captured sporadically or not captured at all.  In the past few years, the computer vision field has found application in many domains by reducing the human burden. Using computer vision systems in the ICU can also potentially enable non-existing assessments or enhance the frequency and accuracy of existing assessments while reducing the staff workload.  In this study, we leverage a state-of-the-art noninvasive computer vision system based on depth imaging to characterize ICU visitations and patients' mobility. We then examine the relationship between visitation and several patient outcomes, such as pain, acuity, and delirium. We found an association between deteriorating patient acuity and the incidence of delirium with increased visitations. In contrast, self-reported pain, reported using the Defense and Veteran Pain Rating Scale (DVPRS), was correlated with decreased visitations. Our findings highlight the feasibility and potential of using noninvasive autonomous systems to monitor ICU patients. 

\end{abstract}

\section{Introduction}

The Intensive Care Unit (ICU) can be a traumatic experience for many patients that adversely affects their long-term physical and cognitive outcomes \cite{halvorsen_patients_2022,cuzco_patients_2022}. Pain \cite{barr_clinical_2013,nordness_current_2021}, immobility \cite{adler_early_2012, tipping_effects_2017}, and sleep deprivation \cite{weinhouse_bench--bedside_2009,boyko_sleep_2012} are common sources of patient distress in the ICU \cite{devlin_clinical_2018}. Prolonged exposure to these causes can induce the onset of delirium \cite{weinhouse_bench--bedside_2009,boyko_sleep_2012}, post-ICU post-traumatic stress disorder (PTSD) \cite{davydow_posttraumatic_2008,righy_prevalence_2019,righy_prevalence_2019}, and increase the mortality risk \cite{pisani_days_2009,ouimet_incidence_2007}.

Monitoring and assessing many aspects of patient care in the ICU is generally performed by nurses or physicians  \cite{barr_clinical_2013}, which can be an error-prone practice \cite{beecroft_sleep_2008,ma_measuring_2017,wang_nurses_2010}. Nurses are often required to work long hours and provide required shift coverage to satisfy patient demand \cite{betsiou_nursing_2022}. These demands often lead to increased stress, sleep deprivation \cite{weaver_sleep_2016,ramadan_association_2014}, and instances of provider burnout \cite{papazian_high-level_2023,jun_relationship_2021,reader_burnout_2008}. Providing care to patients while fatigued can lead to mistakes that negatively affect patient outcomes \cite{barker_effects_2011,weaver_sleep_2016,ramadan_association_2014}. Checking on the patient can also adversely affect the patient's sleep and lead to sleep deprivation \cite{al_mutair_sleep_2020,bihari_factors_2012,eliassen_sleep_2011}.
Patient-care interventions are frequently mentioned as a cause of poor sleep in the ICU \cite{beck_edvardsen_promoting_2020}. Some studies have found that patients rarely go more than 90 minutes without receiving some form of intervention \cite{ritmala-castren_sleep_2015,beck_edvardsen_promoting_2020}. Introducing AI-augmented approaches to monitor patients would reduce providers' workload and incidents of burnout.

Recent advances in AI and sensing technologies have catalyzed the development of AI-augmented solutions to monitor patient mobility and sources of sleep disruption in the critical care setting \cite{gholami_ai_2018}. Ma et al. employed three Microsoft Kinect sensors and the Fast Region-based Convolutional Network (Fast R-CNN) to measure patient activities \cite{ma_measuring_2017,girshick_fast_2015}. Yeung et al. used YOLOv2 to detect and measure the duration of transitional activities, such as getting in or out of bed  \cite{yeung_computer_2019,redmon_yolo9000_2017}. Davoudi et al. used a collection of sensors and deep neural networks to monitor sources of sleep disruption as well as characterize the physical activity and facial expression patterns exhibited by delirious patients \cite{davoudi_intelligent_2019}. Using sound sensors, Davoudi et al. found that average sound pressure levels from the delirious patients’ rooms were roughly equivalent to street traffic noise. This finding is concerning because patients have reported sound as being one of the major etiologies of poor sleep quality \cite{al_mutair_sleep_2020}. 
% Biswal et al. introduced a deep recurrent neural network that predicted patients' sleep stages using electroencephalography (EEG) data\cite{biswal_sleepnet_2017}.
% delirium prediction \cite{davoudi_delirium_2017,ruppert_icu_2020,bandyopadhyay_explainable_2023}, and mobility characterization \cite{ma_measuring_2017, liu_3d_2018,mehra_depth-based_2017}. 

In this study, we examined how patient mobility and personnel-related disturbances correlate with measures of patients' health, i.e., self-reported pain using DVPRS \cite{polomano_psychometric_2016,buckenmaier_iii_preliminary_2013}, patient acuity \cite{shickel_deep_2021,ren_development_2020}, and delirium \cite{ouimet_incidence_2007}.  In an IRB-approved study at the University of Florida (UF), we recruited 44 adult patients admitted to the surgical, medical, trauma, vascular, and neuromedcine ICUs. To quantify patient mobility and sleep disturbances, we employed an Azure Microsoft Kinect sensor \cite{zhang_microsoft_2012} in the ICU to record privacy-preserving depth video \cite{chou_privacy-preserving_2018} of patients during their stay. We developed and deployed a full-stack annotation tool that enabled human annotators to label the presence and posture of individuals in hospital rooms. We used the labeled data to train the YOLOv8 \cite{redmon_you_2016,jocher_yolo_2023} neural network to detect individuals in each depth frame and classify the activity as lying in bed or standing. After training the model on selected annotations, we ran the best-performing model on depth video frame sequences that were temporally aligned with timestamps of clinical outcomes of interest. Within each sequence, we extracted 1) the proportion of lying in bed frames, 2) the average number of people in the room, and 3) the variance of people in the room. These three metrics were examined against our clinical outcomes of interest, i.e., self-reported pain scores, patient acuity \cite{shickel_deep_2021,ren_development_2020}, and delirium. 
 % Given the volume of data (up to 7 days of recording for each patient, 24 hours a day), it is impossible to annotate all the data. Furthermore,  activities such as standing comprise a small percentage of all frames, again adding to the annotation challenges. To improve annotation efficiency and balance diversity, instead of choosing random frames for annotation, we adopted an active learning algorithm that supplied human annotators with the most informative frames for training our model \cite{budd_survey_2021}. 
Our findings revealed significant correlations between indicators of patient outcomes. Patients reported being in less pain when hospital rooms had higher visitations. In contrast, objective measurements, including patient acuity and instances of delirium, were significantly less stable when more individuals were in the room.

% \begin{figure}[H]
% \centering
% \includegraphics[width=\linewidth]{activity recognition.png}
% \caption{\label{fig:main_flowchart}Data capturing and analysis workflow. Depth video and Electronic Health Record (EHR) data were recorded from the ICU. Depth video frames were provided to human annotators using an active learning algorithm designed to increase the diversity of our dataset. The YOLOv8 deep learning model was trained on this dataset and the best-performing model was used to evaluate associations with patients' health metrics.}
% \end{figure}

\section{Methods}

\subsection{Study participants}

The data collected in this study were obtained from adult patients admitted to the surgical, medical, trauma, vascular, and neuromedcine ICUs at the University of Florida Shands Health Hospital. This study was approved by the University of Florida Institutional Review Board by IRB201400546, IRB202101013, and IRB201900354. Informed consent was obtained from all participants, and all procedures complied with the Declaration of Helsinki and university guidelines and regulations.

The cohort used in this study includes 44 patients. Demographic information for the 44 patients is shown in Table \ref{tab:participants}.

\begin{table}[H]
    \centering
\caption{Patients Characteristics}
\vspace*{1mm}
\label{tab:participants}
    \begin{tabular}{lcc}
        \hline
         Characteristics & \multicolumn{2}{c}{Mean} \\
 & N&\% \\
         \hline
         Age&   56.9&15.6\\
 Gender& &\\
             \hspace{5mm} Male&   27& 61\%\\
            \hspace{5mm} Female& 17& 39\%\\
        \hline
         Race &   &\\
         \hspace{5mm} White &   36& 82\%\\
         \hspace{5mm} Black &   5&11\%\\
\hspace{5mm} Asian & 2& 5\%\\
\hspace{5mm} Other & 1& 2\%\\
        \hline
 Hospital stay \\(hours) & 93.6& \\
        \hline
 Outcomes & & \\
\hspace{5mm} Pain score & 2.4 &  \\
\hspace{5mm} Acuity & 0.69 & \\
\hspace{5mm} Delirium & 0.24 & \\
        \hline

    \end{tabular}
\end{table}

\subsection{Data Collection}
We captured depth frames in the ICU using Microsoft Kinect cameras \cite{zhang_microsoft_2012}, which captured 3D volumetric data of the ICU environment while safeguarding the physical identities of each person. Six compact capture and compute systems were used for data collection, each equipped with a depth camera and a compact computer. The cameras had a field of view (FOV) of $70^\circ$ x $60^\circ$ and produced 640 x 576 px image frames. Frames were saved at a frequency of 1 FPS. Clinical coordinators oversaw the deployment and use of these systems to ensure patient care was not impeded.

 After the data was captured by the local capture and compute system, it was encrypted, compressed, and transferred through a secure link to a high-performance server for data annotation and model training. 
 
 \subsection{Data Annotation}
\begin{figure}[H]
\centering
\includegraphics[width=1\linewidth]{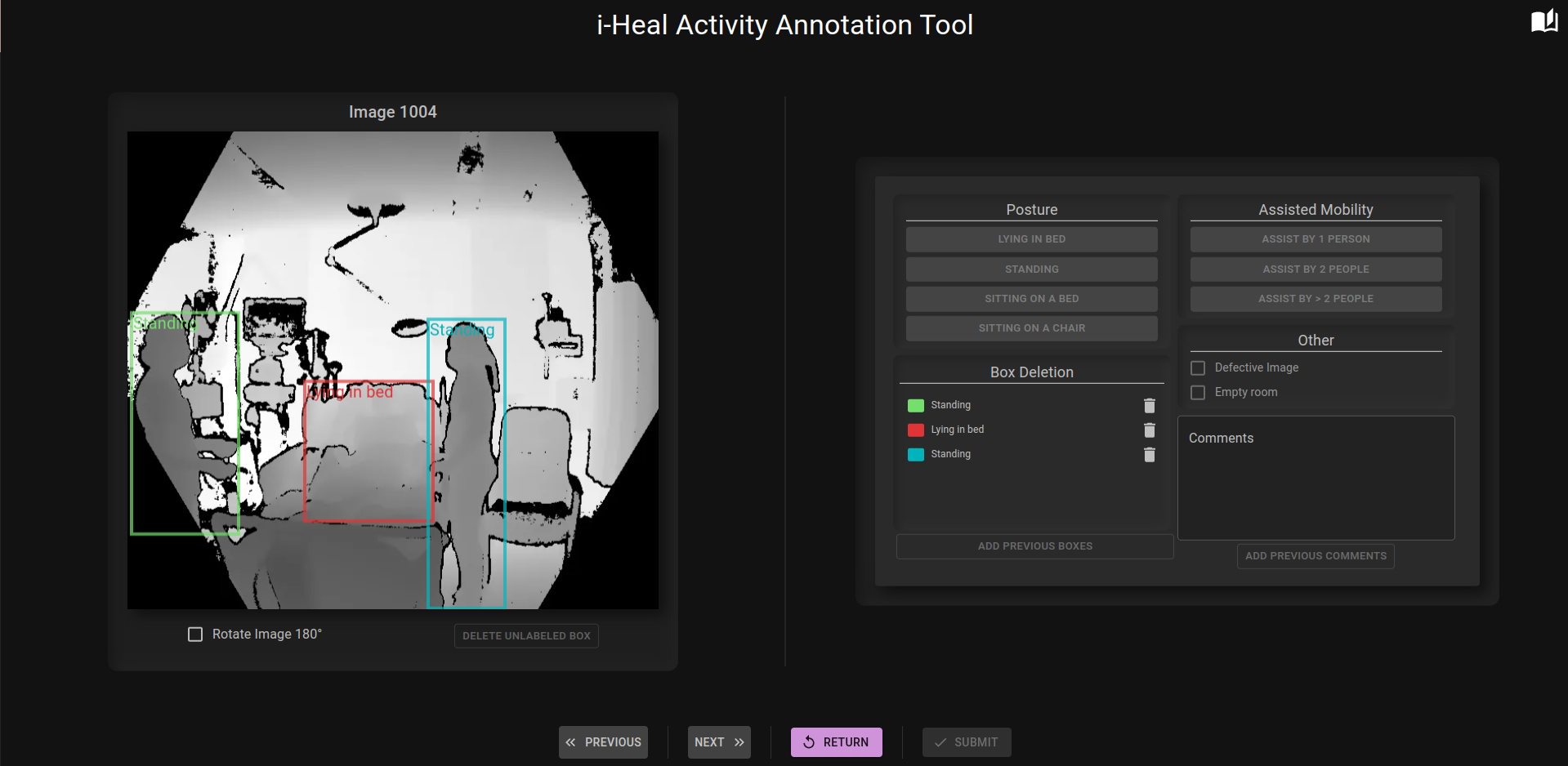}
\caption{\label{fig:annotation_tool}The front-end user interface to annotate depth frames. Posture classes include lying in bed, standing, sitting on a bed, and sitting on a chair. The tool also includes an assisted mobility class, which is intended to capture instances when someone is physically assisting the patient with their mobility.}
\end{figure}

 To build our mobility detection dataset, we decided to develop an in-house bounding box annotation tool that would enable remote users to complete annotations without installing any software. We designed our depth annotation tool as a full-stack web application that enables annotators to draw bounding boxes around every person portrayed in an image and label each box as one of the following postures: lying in bed, standing, sitting on a chair, and sitting on a bed. The front-end user interface was developed using the React JavaScript framework (Figure \ref{fig:annotation_tool}). The front end was connected to a back-end MongoDB database using the FastAPI Python framework. Before starting the annotation, annotators completed the necessary steps to acquire Institutional Review Board (IRB) approval. They were provided with a Standard Operating Procedures (SOP) document that comprehensively outlined the correct annotation procedures for each frame, and they were assigned an initial training dataset.  An additional annotation reviewer evaluated the quality of the annotations.

\subsection{YOLOv8 Architecture}

We used the YOLOv8 \cite{terven_comprehensive_2023} architecture for our patient activity detection model. The YOLOv8's backbone is the Darknet-53 convolutional neural network with an added C2f module \cite{terven_comprehensive_2023,redmon_yolov3_2018}. The C2f module, which stands for "cross-stage partial bottleneck with two convolutions," improves detection accuracy by combining high-level features with contextual information. To improve YOLOv8's ability to detect smaller objects, bounding box loss is computed using the complete intersection over union (CIoU) \cite{zheng_distance-iou_2020} and distribution focal loss (DFL) \cite{li_generalized_2020} loss functions. The YOLOv8 architecture also integrates feature pyramid networks, which are adept at recognizing objects at different scales. This is beneficial for our depth frame detection task, given the varying distances of patients within the frames. 

\subsection{Training Pipeline}

% We adopted an active learning approach to select which samples were assigned to annotators. First, our model was trained on an initial batch of data. The trained model was then used to query the next round of annotations to send to annotators. To reduce the large number of frames that included a single person lying in bed (the majority of the frames, given the critical condition of most patients), we ranked frames based on the number of individuals in the room. This process was repeated to obtain the model used in the present study. 
The annotated dataset in this study included 30,840 annotated depth frames from 44 patients. Model performance was evaluated using 5-fold cross-validation. Patient IDs were used to split training and test sets within each fold to prevent data leakage \cite{lekadir_future-ai_2021}. In each fold, 20\% of the patients were allocated for testing, while the remaining patients were used for training. Experiments were conducted on a single A100 GPU. The default hyperparameters provided by Ultralytics \cite{jocher_yolo_2023} were used for our experiments. The model was trained for 100 epochs using stochastic gradient descent (SGD) and a learning rate of 0.01. We report the mean average precision (mAP), precision, recall, and F1-Score using an intersection over union threshold (IoU) of 0.5 as an average across five folds.
% and three experiments with unique randomization seeds

\subsection{Characterizing mobility and sleep disturbances}

The total number of detected \textit{lying in bed} and \textit{standing} labels within each frame was used to measure patient visitations average and variance. 
Walking in and out of rooms and checking different equipment can produce noise and disturb patients' sleep. We evaluated sleep disturbances during both day and night as it has been shown that approximately 50\% of ICU patients' sleep occurs during the day \cite{al_mutair_sleep_2020,beck_edvardsen_promoting_2020,closs_patients_1992}.

% Since we employed depth video frames for patient monitoring, it was challenging to identify which individual in the room was the patient. To address this problem, we relied on a heuristic solution. Each hospital room within this cohort featured a single, large bed. We determined that if someone was lying in bed, they were likely the patient. Using this assumption, patient immobility was characterized as the proportion of frames that someone was lying in bed.

\subsection{Patient Outcomes}
We assessed patient outcomes by evaluating three distinct scores, namely self-reported pain, patient acuity \cite{shickel_deep_2021,ren_development_2020}, and occurrences of delirium. The patients reported pain using the Defense and Veterans Pain Rating Scale (DVPRS) \cite{buckenmaier_iii_preliminary_2013}, which is a numerical scale ranging from 0 to 10, where 0 indicated the absence of pain, 1-4 signified mild pain, 5-6 indicated moderate pain, and 7-10 represented severe pain. Patient acuity was evaluated using the methodology proposed by a recent study \cite{shickel_deep_2021,ren_development_2020}. Patient acuity was classified into two states: stable and unstable. Patients were classified as unstable if, during the specified time interval, they received continuous renal replacement therapy (CRRT), required ventilation, received a vasopressor infusion, or required a blood transfusion of 10 units or more in the past 24 hours. Delirium was evaluated using the Richmond Agitation Sedation Scale (RASS), the Confusion Assessment Method for the Intensive Care Unit (CAM-ICU), and the Glasgow Coma Scale (GCS) \cite{reznik_fluctuations_2020,jain_glasgow_2018}.

\subsection{Associating model metrics with patient Outcomes}
The timestamps corresponding to clinical events of interest were used to select temporally aligned depth video frames. Patient acuity was recorded in four-hour intervals. For pain record timestamps, video frames within the 15-30 minute period prior to the pain timestamp were selected. We excluded the 15-minute window preceding each timestamp as it was likely to encompass the frames captured when the caregiver was in the room. Including these frames has the potential to introduce bias. For delirium record timestamps, video frames were chosen if they occurred in the 15-60 minute window preceding the recorded time.

Model metrics were recorded for each respective sequence of video frames, then evaluated for normality using D'Agostino and Pearson's test \cite{dagostino_tests_1973,dagostino_omnibus_1971}. Because the null hypothesis that each sample came from a normal distribution could be rejected, the Mann-Whitney U rank test \cite{mann_test_1947} was used to evaluate independence between samples. Samples were also split based on whether the record timestamps occurred during the day (7 AM - 7 PM) or night (7 PM - 7 AM).
(need to mention type 1 error correction)
\section{Results}

\subsection{YOLOv8 experiments}

In this section, we report the performance of the YOLOv8 models that were trained on our mobility detection dataset and evaluated using 5-fold cross-validation. In Table \ref{tab:yolov8_results}, we present the experimental results of our posture detection models across five independent test sets and report average mAP, Precision, Recall, and F1 Score. The results are stratified by the predicted posture. The model obtained a mAP of 0.88 for the standing class and a mAP of 0.79 for the lying-in-bed class.

\begin{table}[H]
    \centering
\caption{Experimental results on test dataset}
\label{tab:yolov8_results}
    \begin{tabular}{ccccc}
    \hline
 & \multicolumn{4}{c}{Performance}\\
    
 Class& \multicolumn{4}{c}{Mean (STD)}\\
        % \hline
         &  mAP&  Precision&  Recall& F1 Score\\
         \hline
         Lying in bed&  0.79 (0.21)&  0.79 (0.20)&  0.74 (0.18)& 0.77 (0.19)\\
         Standing&  0.88 (0.17)&  0.90 (0.03)&  0.83 (0.21)& 0.84 (0.14)\\
         All classes&  0.83 (0.17)&  0.84 (0.09)&  0.79 (0.18)& 0.81 (0.14)\\
         \hline
    \end{tabular}
    
\end{table}

\subsection{Associations with Patient Outcomes}\label{AssocPatientOutcomes}
For ICU room visitations, our study uncovered noteworthy positive correlations with subjective measures of patient outcomes, such as self-reported pain, alongside negative correlations with objective measures like delirium and acuity. This relationship persisted even when we divided the samples into night and day occurrences, revealing consistent patterns across all periods. The following sections explore each patient metric in more detail. 

\subsubsection{Self-Reported Pain}
Patients consistently reported lower pain scores when more people were in the room (Table \ref{tab:pain_correlations}). In table \ref{tab:pain_correlations}, pain groups were combined based on whether the pain scores were classified into the no pain and mild pain groups versus the moderate and severe pain groups. The Kruskal-Wallis H-test \cite{kruskal_use_1952} was also used to compare the four pain groups separately, but no significant difference was found ($p = 0.9$). The proportion of frames with the patient lying in bed was higher for the no and mild pain group compared to the moderate and severe pain group, but the difference wasn't statistically significant. 

\begin{figure}[H]
\centering
\includegraphics[width=0.75\linewidth]{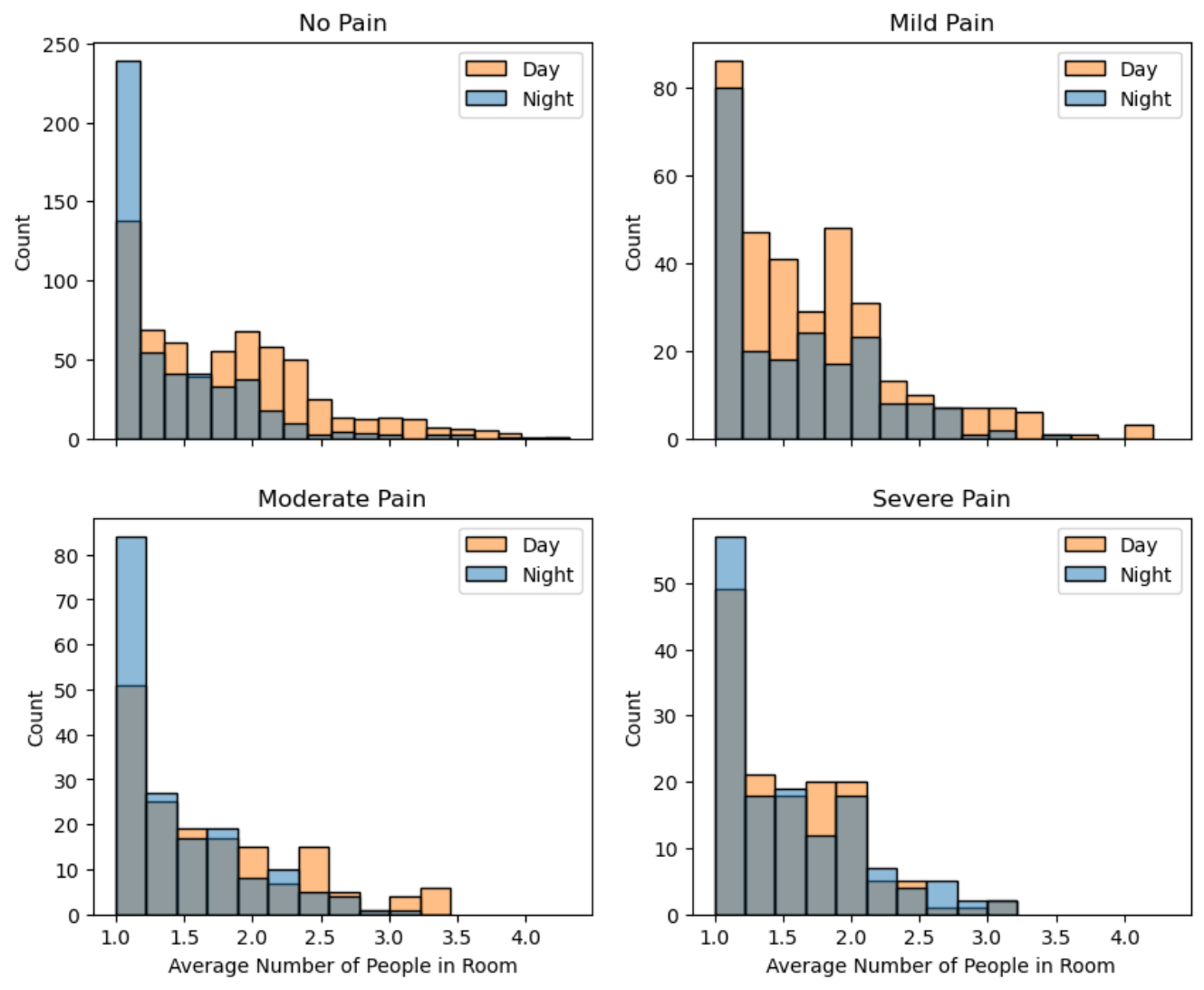}
\caption{\label{fig:pain_histograms}Relationship between pain levels and the average number of people in the room during the night versus day.}
\end{figure}

Splitting samples based on whether they occurred during the day versus night yielded similar trends as those found with the combined datasets (Table \ref{tab:pain_correlations}). The difference between night and day samples for the average number of people detected is shown in  Figure \ref{fig:pain_histograms}.
 % Another interesting finding was that the difference between hospital room visitation average during the night versus day was more significant for no and mild samples (Day Average $= 0.61$, Night Average $=0.26$) than the moderate and severe pain sequences. 

\begin{table}[H]
    \centering
\caption{Pain Associations with Model Metrics}
\label{tab:pain_correlations}
\vspace*{1mm}
  % \centering
    \begin{tabular}{m{3cm}|ccc|ccc|c}
    \hline 
         \multirow{2}{*}{Model Metrics}  &\multicolumn{3}{c|}{No/Mild pain}&\multicolumn{3}{c|}{Moderate/High Pain} &  \multirow{2}{*}{p-value}\\
              % &&& \multicolumn{2}{c}{(n=333)} &  \multicolumn{2}{c}{(n=257)} &   \\[1ex]
         &N&  Mean (STD)&Max (Min)&N&Mean (STD) & Max (Min) &   \\
    \hline
\raggedright{Visitation Average}&&&&&&& \\

\hspace{\tabindent}{Day}&{973}&{1.77 (0.65)}&{4.32 (1.0)}&{307}&{1.62 (0.58)}&{3.45 (1.0)}&{\boldmath$3.06e^{-04}$}\\ 
      \hspace{\tabindent}{Night}&{699}&{1.45 (0.52)}&{4.31 (1.0)}&{320}&{1.48 (0.51)}&{3.22 (1.0)}&{0.219}\\
      \hspace{\tabindent}{Combined}&{1672}&{1.64 (0.62)}&{4.32 (1.0)}&{627}&{1.55 (0.55)}&{3.45 (1.0)}&{\textbf{0.008}}\\[2ex]
 \raggedright{Visitation Variance}&&&&&&& \\
        \hspace{\tabindent}{Day}&{973}&{0.41 (0.4)}&{3.0 (0.0)}&{307}&{0.34 (0.34)}&{2.32 (0.0)}&{\textbf{0.002}}\\ 
      \hspace{\tabindent}{Night}&{699}&{0.24 (0.28)}&{1.88 (0.0)}&{320}&{0.26 (0.31)}&{2.27 (0.0)}&{0.402}\\
      \hspace{\tabindent}{Combined}&{1672}&{0.34 (0.36)}&{3.0 (0.0)}&{627}&{0.3 (0.33)}&{2.32 (0.0)}&{\textbf{0.01}}\\[1ex]
    \raggedright{Lying in Bed Proportion}&&&&&&& \\
          \hspace{\tabindent}{Day}&{973}&{0.89 (0.26)}&{1.0 (0.0)}&{307}&{0.87 (0.29)}&{1.0 (0.0)}&{0.305}\\ 
      \hspace{\tabindent}{Night}&{699}&{0.95 (0.19)}&{1.0 (0.0)}&{320}&{0.91 (0.26)}&{1.0 (0.0)}&{0.093}\\
      \hspace{\tabindent}{Combined}&{1672}&{0.91 (0.23)}&{1.0 (0.0)}&{627}&{0.89 (0.27)}&{1.0 (0.0)}&{0.549}\\[1ex]
\hline
\end{tabular}
\end{table}

\subsubsection{Patient Acuity}
\begin{figure}[H]
\centering
\includegraphics[width=0.75\linewidth]{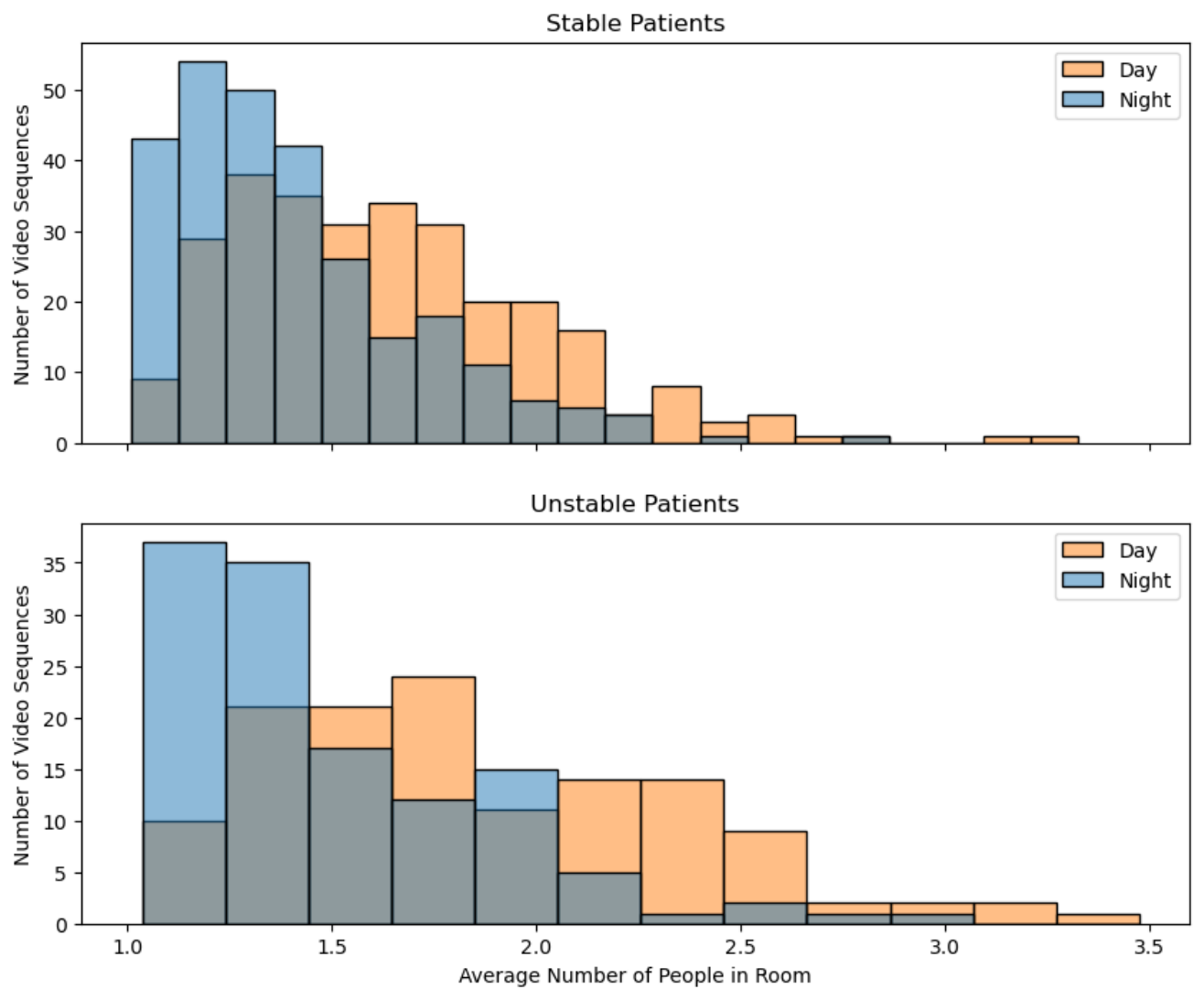}
\caption{\label{fig:acuity_histograms}Relationship between patient acuity and the average number of people in the room during the night versus day.}
\end{figure}

For patient acuity, hospital room visitations average and variance were both significantly higher for unstable frame sequences compared to stable sequences (Table \ref{tab:acuity_model_metrics}). The same pattern was found for the proportion of lying-in-bed samples, but the difference between groups wasn't statistically significant. Similar to the patterns found for the pain associations, average day versus night hospital room visitations and variance were greater for stable patients than unstable patients (Table \ref{tab:acuity_model_metrics}). The difference between night and day samples for the average number of people detected is shown in  Figure \ref{fig:acuity_histograms}.

\begin{table}[H]
    \centering
\caption{Patient Acuity Associations with Model Metrics}
\vspace*{1mm}
\label{tab:acuity_model_metrics}
     \begin{tabular}{m{3cm}|ccc|ccc|c}
    \hline 
         \multirow{2}{*}{Model Metrics}  &\multicolumn{3}{c|}{Stable}&\multicolumn{3}{c|}{Unstable} &  \multirow{2}{*}{p-value}\\
              % &&& \multicolumn{2}{c}{(n=333)} &  \multicolumn{2}{c}{(n=257)} &   \\[1ex]
         &N&  Mean (STD)&Max (Min)&N&Mean (STD) & Max (Min) &   \\
    \hline
     \raggedright{Visitation Average}&&&&&&& \\
                 \hspace{\tabindent}{Day}&{294}&{1.64 (0.38)}&{3.33 (1.04)}&{133}&{1.85 (0.5)}&{3.48 (1.04)}&{\boldmath$5.05e^{-05}$}\\ 
      \hspace{\tabindent}{Night}&{285}&{1.41 (0.3)}&{2.83 (1.01)}&{127}&{1.5 (0.38)}&{2.93 (1.05)}&{\textbf{0.034}}\\
      \hspace{\tabindent}{Combined}&{579}&{1.52 (0.36)}&{3.33 (1.01)}&{260}&{1.68 (0.48)}&{3.48 (1.04)}&{\boldmath$3.96e^{-05}$}\\[1ex]
 \raggedright{Visitation Variance}&&&&&&& \\
                 \hspace{\tabindent}{Day}&{294}&{0.56 (0.31)}&{1.95 (0.05)}&{133}&{0.68 (0.38)}&{2.48 (0.05)}&{\textbf{0.001}}\\ 
      \hspace{\tabindent}{Night}&{285}&{0.38 (0.27)}&{1.98 (0.01)}&{127}&{0.42 (0.27)}&{1.77 (0.05)}&{\textbf{0.035}}\\
      \hspace{\tabindent}{Combined}&{579}&{0.47 (0.31)}&{1.98 (0.01)}&{260}&{0.56 (0.36)}&{2.48 (0.05)}&{\boldmath$2.51e^{-04}$}\\[1ex]
    \raggedright{Lying in Bed Proportion}&&&&&&& \\
          \hspace{\tabindent}{Day}&{294}&{0.85 (0.25)}&{1.0 (0.0)}&{133}&{0.93 (0.14)}&{1.0 (0.11)}&{0.082}\\ 
      \hspace{\tabindent}{Night}&{285}&{0.93 (0.2)}&{1.0 (0.0)}&{127}&{0.96 (0.13)}&{1.0 (0.01)}&{0.64}\\
      \hspace{\tabindent}{Combined}&{579}&{0.89 (0.23)}&{1.0 (0.0)}&{260}&{0.95 (0.13)}&{1.0 (0.01)}&{0.161}\\[1ex]
\hline
\end{tabular}
\end{table}

\subsubsection{Delirium}
 Associations with delirium exhibited similar trends to those found for patient acuity. Hospital room visitation average and variance were significantly higher for delirium frame sequences than non-delirious sequences (Table \ref{tab:delirium_model_metrics}). The same trend was found for the proportion of lying-in-bed samples, but the difference between groups wasn't statistically significant. Differences between day versus night hospital room visitation average and variance were more significant for delirium samples than for non-delirious samples. Comparing day versus night differences for delirium samples isn't as reliable because the sample sizes were smaller compared to those for pain and acuity.

\begin{table}[H]
    \centering
\caption{Delirium Associations with Model Metrics}
\vspace*{1mm}
\label{tab:delirium_model_metrics}
  \begin{tabular}{m{3cm}|ccc|ccc|c}
    \hline 
         \multirow{2}{*}{Model Metrics}  &\multicolumn{3}{c|}{Non-Delirious}&\multicolumn{3}{c|}{Delirium} &  \multirow{2}{*}{p-value}\\
              % &&& \multicolumn{2}{c}{(n=333)} &  \multicolumn{2}{c}{(n=257)} &   \\[1ex]
         &N&  Mean (STD)&Max (Min)&N&Mean (STD) & Max (Min) &   \\
    \hline
     \raggedright{Visitation Average}&&&&&&& \\
    \hspace{\tabindent}{Day}&{79}&{1.51 (0.42)}&{2.51 (1.0)}&{26}&{2.06 (0.53)}&{2.98 (1.01)}&{\boldmath$7.99e^{-06}$}\\ 
      \hspace{\tabindent}{Night}&{87}&{1.6 (0.59)}&{4.43 (1.0)}&{28}&{1.48 (0.51)}&{2.83 (1.0)}&{0.245}\\
      \hspace{\tabindent}{Combined}&{166}&{1.56 (0.52)}&{4.43 (1.0)}&{54}&{1.76 (0.59)}&{2.98 (1.0)}&{\textbf{0.038}}\\[1ex]
 \raggedright{Visitation Variance}&&&&&&& \\
      \hspace{\tabindent}{Day}&{79}&{0.35 (0.31)}&{1.44 (0.0)}&{26}&{0.58 (0.41)}&{2.01 (0.01)}&{\textbf{0.001}}\\ 
      \hspace{\tabindent}{Night}&{87}&{0.34 (0.28)}&{1.43 (0.0)}&{28}&{0.36 (0.4)}&{1.34 (0.0)}&{0.669}\\
      \hspace{\tabindent}{Combined}&{166}&{0.34 (0.29)}&{1.44 (0.0)}&{54}&{0.47 (0.42)}&{2.01 (0.0)}&{\textbf{0.047}}\\[1ex]
    \raggedright{Lying in Bed Proportion}&&&&&&& \\
          \hspace{\tabindent}{Day}&{79}&{0.96 (0.13)}&{1.0 (0.0)}&{26}&{0.93 (0.16)}&{1.0 (0.32)}&{0.389}\\ 
      \hspace{\tabindent}{Night}&{87}&{0.93 (0.22)}&{1.0 (0.0)}&{28}&{0.99 (0.03)}&{1.0 (0.89)}&{0.573}\\
      \hspace{\tabindent}{Combined}&{166}&{0.94 (0.18)}&{1.0 (0.0)}&{54}&{0.96 (0.11)}&{1.0 (0.32)}&{0.846}\\[1ex]
\hline
\end{tabular}
   %  \begin{tabular}{|c||cc|cc|c|}
   %  \hline
   %       % &\multicolumn{2}{c|}{Number of people in room}&\multicolumn{2}{c|}{Number of people in room}\\
   %       \multirow{3}{*}{Model Metrics} &\multicolumn{2}{c|}{Non-Delirious Sequences}&\multicolumn{2}{c|}{Delirium Sequences} &  \multirow{3}{*}{p-value}\\
   %          % &&\multirow{3}{c|}&&&&&&\\
   %            & \multicolumn{2}{c|}{(n=166)} &  \multicolumn{2}{c|}{(n=54)} &   \\[0.5ex]
   %       &  Mean (STD) & Max (Min) &  Mean (STD) & Max (Min) &   \\
   %      \hline
   %     {Number of People}&\multirow{2}{*}{1.56 (0.52)}
   %  &\multirow{2}{*}{4.43 (1.0)}
   %  &\multirow{2}{*}{1.76 (0.59)}
   %  &\multirow{2}{*}{2.98 (1.0)}
   %  &\multirow{2}{*}{0.038}\\
   %  Average &&&&&\\
   %  \hline
   %  {Number of People}
   %  &\multirow{2}{*}{0.34 (0.29)}
   %  &\multirow{2}{*}{1.44 (0.0)}
   %  &\multirow{2}{*}{0.47 (0.42)}
   %  &\multirow{2}{*}{2.01 (0.0)}
   %  &\multirow{2}{*}{0.047}\\
   %  Variance &&&&&\\
   %  \hline
   %   {Lying in Bed}
   % &\multirow{2}{*}{0.94 (0.18)}
   %  &\multirow{2}{*}{1.0 (0.0)}
   %  &\multirow{2}{*}{0.96 (0.11)}
   %  &\multirow{2}{*}{1.0 (0.32)}
   %  &\multirow{2}{*}{0.846}\\
   %  Proportion &&&&&\\
   %  \hline
   %  \end{tabular}

\end{table}

\section{Discussion}
In this study, we employed the state-of-the-art YOLOv8 deep learning model \cite{jocher_yolo_2023} to detect patient postures. The results demonstrated that the model had a harder time detecting the "\textit{lying in bed} class compared to the \textit{standing} class. This result aligns with our expectations because individuals lying in bed were mostly covered by thick covers that made it difficult to distinguish which features belonged to the person. 
% Integrating an active learning pipeline to determine the data sent to annotators led to a notable enhancement in the training dataset's diversity and the annotations' efficiency, which also resulted in an overall improvement in model performance.

The metrics extracted from the model were the proportion of \textit{lying in bed} frames and the average and variance of hospital room visitations. The proportion of \textit{lying in bed} frames was evaluated because we could assume that if someone was lying in bed, that person was the patient. If every frame in a sequence contained a person lying in bed, that would indicate that they were immobile.

Among the metrics extracted from the model, hospital room visitations average and variance exhibited substantial correlations with measures of patient outcomes. These findings are logical because noise is a crucial factor associated with sleep deprivation and the onset of delirium \cite{mistraletti_sleep_2008,cupka_effect_2020}, and each individual represents a unique source of noise. This observation highlights the importance of using noninvasive autonomous systems to monitor patients. Individuals who tend to wake easily during sleep may be disturbed when healthcare providers open doors or inspect equipment  \cite{al_mutair_sleep_2020,bihari_factors_2012,eliassen_sleep_2011}. Minimizing the need for healthcare providers to enter or exit hospital rooms would allow patients to experience uninterrupted sleep and facilitate their recovery. Furthermore, implementing a noninvasive system for tracking patient mobility would allow care providers to remotely assess whether a patient is asleep and delay room entry until the patient is awake.

Another intriguing discovery was the difference in the relationships identified when comparing subjective and objective measures of patient outcomes. Higher levels of hospital room visitations were linked to adverse objective measures of patient outcomes, including patient acuity and delirium. Conversely, self-reported pain exhibited a contrasting pattern. According to Boring et al., patients tend to downplay their pain when in the presence of friends or family members, as they wish to avoid causing undue worry \cite{boring_how_2021}. Since we couldn't identify anyone in each frame, it wasn't possible to know whether the people in the room were care providers or family members. Incorporating each person's relationship to the patient in video sequences would make it possible to explore the impact of family presence on self-reported pain records further.

While a negative correlation was observed between patient outcomes and the number of people in each room, the current analysis does not provide adequate evidence to establish a causal relationship. The patient might have pressed the nurse call button \cite{galinato_perspectives_2015} to request aid or their condition may have worsened to the extent that emergency medical staff were called in by the patient's monitoring equipment. This would suggest that the number of individuals in the room resulted from the patients' compromised outcomes. Incorporating the timing of each event that could potentially disturb a patient is needed to evaluate whether there is a causal relationship between hospital room visitations and patients' outcomes. 

Using the average and variance of the number of people in an ICU room only considers one factor that can adversely affect patients' sleep. Other stimuli, such as light or other sources of noise, have also been shown to impact patients' sleeping habits \cite{al_mutair_sleep_2020,bihari_factors_2012,eliassen_sleep_2011}. Incorporating real-time data from noise and light sensors would improve our system's capacity to identify the critical factors contributing to sleep deprivation and notify care providers of potentially serious events, such as the onset of delirium.

Monitoring the average amount of time a patient was lying in bed indicates how much they were resting, but it doesn't reveal how much time they spent sleeping. Combining depth video with other sensors, such as a heart rate monitor, an RGB video camera, or a polysomnography device, would improve our system's ability to monitor sleeping patterns. 

There are advantages and disadvantages associated with the utilization of depth cameras within an ICU as opposed to RGB cameras. Depth cameras provide 3D information about a scene and preserve the privacy of each person in a scene as long as the resolution is low. However, identifying which person is the patient can be a challenging endeavor when employing a depth camera. In the future, we will incorporate patient tracking using one of the object tracking architectures that have been integrated into the YOLOv8 model \cite{zhang_bytetrack_2022,aharon_bot-sort_2022}. Using tracking information, we can follow the person lying in the hospital bed and evaluate their posture whenever they move around the hospital room. Knowing which person the patient is will enable better mobility characterization throughout the patient's recovery period in the ICU. We can establish a comprehensive mobility index by amalgamating this data with information from other modalities, such as accelerometer data. This index would be valuable for preventing adverse events and enhancing patient outcomes.

\section{Conclusions}
 
We developed a full-stack application that enabled annotators to label 30,840 depth frames across 44 patients recovering in an ICU. We used this dataset to train the state-of-the-art YOLOv8 deep learning model to characterize patient mobility and evaluate hospital room visitations.

Our trained model revealed significant relationships between indicators of patient outcomes and hospital room visitations. Subjective assessments, such as self-reported pain, were lower when rooms had higher visitations. In contrast, objective measurements, including patient acuity and instances of delirium, were significantly higher when more individuals were in the room. These findings illustrate the following: (1) the number of people in a hospital room can be a potential source of bias for self-reported pain scores; (2) greater quantities of people in a hospital room may have adverse effects on the patient, such as disturbing the patients' sleep.

The dataset utilized in this report is still in development and will be expanded to incorporate patient tracking, enabling granular characterization of their mobility and the identification of instances where patients' mobility is being assisted by other care providers. We aim to integrate this system into an intelligent suite of sensors backed by state-of-the-art AI technologies, enabling real-time monitoring and analyses of patient outcomes in the ICU. By further refining and expanding our dataset, we aim to enhance the capabilities of our system to detect and respond to changes in patient mobility and overall health. This integrated approach holds the potential to revolutionize patient care, offering timely interventions, reducing human error, and ultimately improving the quality of care provided in intensive care units.
\section{Acknowledgements}
A.B, P.R., and T.O.B. were supported by NIH/NINDS R01 NS120924, NIH/NIBIB R01 EB029699. P.R. was also supported by NSF CAREER 1750192.

\nocite{openai_chatgpt_2021}
\printbibliography
\end{document}